\documentclass[11pt]{article}
\usepackage{geometry}
\usepackage{coling2020}
\usepackage{times}
\usepackage{url}
\usepackage{latexsym}
\usepackage{microtype}

\usepackage{graphicx}
\usepackage{wrapfig}
\usepackage{amssymb}
\usepackage{amsmath}
\usepackage{textcomp}
\usepackage{hyperref}
\usepackage{subcaption}

\usepackage{caption}
\colingfinalcopy 

\title{HCMS at SemEval-2020 Task 9: A Neural Approach to Sentiment Analysis for Code-Mixed Texts}

\author{Aditya Srivastava \\
  International Institute of \\Information Technology, \\Hyderabad \\
  {\tt aditya.srivastava}\\{\tt @research.iiit.ac.in} \\\And
  V. Harsha Vardhan \\
  International Institute of \\Information Technology, \\Hyderabad \\
  {\tt harshavardhan.v}\\{\tt @research.iiit.ac.in} \\}

\date{}

\begin{document}
\maketitle
\begin{abstract}
  Problems involving code-mixed language are often plagued by a lack of resources and an absence of materials to perform sophisticated transfer learning with. In this paper we describe our submission to the Sentimix Hindi-English task involving sentiment classification of code-mixed texts, and with an F1 score of 67.1\%, we demonstrate that simple convolution and attention may well produce reasonable results.
\end{abstract}

\section{Introduction}
\label{intro}

%
%
\blfootnote{
    %
    %
    %
    %
    \hspace{-0.65cm}  
    This work is licensed under a Creative Commons 
    Attribution 4.0 International Licence.
    Licence details:
    \url{http://creativecommons.org/licenses/by/4.0/}.
    %
    %
}

Sentiment analysis has gained importance in the current world where social media is crucial in gauging public opinion on products, political campaigns, latest trends and more. A large portion of the world's population is multilingual, and so is the content on social networks. Code-mixing is a natural phenomenon among bilinguals where phrases and words from one language are employed in another. Typically, the underlying grammar of the primary language that is being spoken is kept intact and phrases from another language are embedded into it. India in particular, has 23 officially recognized languages and a majorly bilingual population, requiring robust computational tools to effectively exploit the data it produces.

Deep neural networks have had great success in predicting sentiment encapsulated in text - a big part of this success has been their ability to utilize pretrained word embeddings. In code-mixed tweets however, Hindi words are written in Roman script instead of its native Devanagari script, forcing one to use a transliterator in order to take advantage of pretrained aligned word embeddings, like \textit{fastText} \cite{bojanowski2016enriching}. These words in Roman script are unnormalized and may have multiple spellings; for example, \textit{bahut} can be spelt in Roman script as \textit{bahut}, \textit{bohot}, \textit{bohut} etc., and such errors propagate from the transliterator to our downstream task, thereby reducing performance.
 
In this work, we propose a deep convolution network with self-attention which shows promising results, without any pretraining. Convolution neural networks (CNNs) are known to capture local relations \cite{kim2014convolutional} or \textit{local context}, and work here as feature extractors acting as a substitute for handcrafted sentiment features. A self-attention layer is then applied over these features (presenting as vectors), which allow each individual feature to attend to all other features \cite{bahdanau2014neural}, providing \textit{global context}. We call this the \textbf{Hierarchical Context Modeling System} (henceforth referred to as \textbf{HCMS}), due to the hierarchical nature of context extraction performed on both levels.

\section{Background}
\label{background}

Early sentiment analysis systems made use of corpus based statistics \cite{wiebe2000learning}, linguistic tools like Wordnet \cite{miller1998wordnet} and lexicon based classifiers \cite{turney2002unsupervised}. More recently, a large portion of work has involved Recurrent Neural Networks (RNNs), Long Short Term Memory Networks (LSTMs) and CNNs at word and character levels  \cite{DEMULDER201561,zhang2015characterlevel}. Usage of sub-word level representations (based on character n-grams) \cite{joshi2016towards}, combining both CNNs and LSTMs have also been demonstrated, where the CNN was used to generate sub-word features which were then augmented to word level features. Sharma et al. \shortcite{7275819} also set precedent for normalizing code mixed text by dealing with the multiple transliterated word forms being generated as a result of the Romanization of non-English scripts.

Attention mechanisms were first introduced for neural machine translation, and achieved state-of-the-art results at the time. The idea, which produces significant gains in our own model, has been applied to a variety of tasks such as image captioning, text classification, question answering \cite{Cheng_2016,xu2015attend,choi-etal-2017-coarse}, and even aspect level sentiment classification for more fine grained feature extraction \cite{wang2016attention}. Attention helps in obtaining better representations of text which in turn boosts downstream performance and thus, is a vital component of HCMS where self-attention is used to boost the performance of the underlying architecture. 

\section{System Overview}
\label{system_overview}

\begin{wrapfigure} {r}{0.4\textwidth}
    \vspace{-30pt}
    \begin{center}
        \includegraphics[page=1, trim = 30mm 20.71cm 14.46cm 11mm, clip, width=0.25\textwidth]{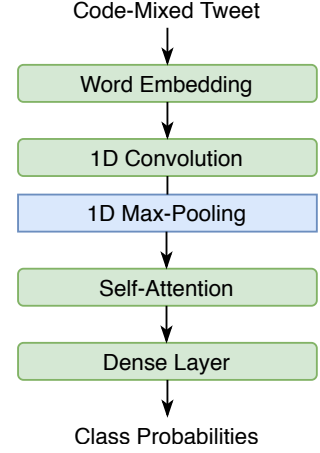}
    \end{center}
    \vspace{-5pt}
    \caption{Model Architecture}
    \label{fig:model_architecture}
\end{wrapfigure}


Approaching the problem as one of sequence classification, we present the HCMS architecture as depicted in  figure \ref{fig:model_architecture}. For every tweet (represented as a sequence of word embeddings), our model first performs a convolution and max-pooling operation to get local context between words that are adjacent to one another. In the next step, self-attention allows every context vector to attend to every other context vector in the sentence to get global context, summarizing the tweet. Lastly, the model passes this summary through a dense layer to produce probabilities over the three classes, namely positive, negative and neutral.

\subsection{Algorithm}
Given a tweet $X$ as $[x_1, ..., x_u]$, where every $x_i\in\mathbb{R}^d$  is a $d$ dimensional word embedding. The model first performs a ReLU activated, 1D convolution operation along the tweet  followed by a max-pooling operation, to produce localized context vectors $C$ as $[c_1, ...c_v]$, where every $c_i\in\mathbb{R}^{d'}$. The dimensions of $C$ will depend on the parameters of the CNN and max-pooling layers.

\begin{equation}
    C = \text{MaxPool}(\text{ReLU}(\text{CNN}(X))
\end{equation}

Next, the model performs self-attention between the local context vectors in $C$. Self-attention lets every $c_i\in C$ attend to every other $c_{i'}\in C$ to produce attention vectors $A$ as $[a_1, ...a_v]$. These vectors are then concatenated to produce global context vector $G$ as $[g_1, ...g_w]$, where every $g_i\in\mathbb{R}$.

\vspace{-10pt}
\begin{align}
\text{For every $t^\text{th}$ local context vector,} \nonumber \\
    h_{t,t'} &= \text{tanh}(c^T_tW_t + c^T_{t'}W_c + b_t) \nonumber && \text{$c_t, c_{t'}\in C$ and $t\neq t'$}\\
    e_{t,t'} &= \sigma(W_ah_{t,t'} + b_a) \nonumber \\
    q_t &= \text{Softmax}(e_t) \nonumber \\
    a_t &= \sum_{t'} q_{t,t'}c_t \nonumber  && \text{$a_t \in A$} \\
    G &= \text{Concatenate}(l)
\end{align}

Finally, the global context vector $G$ is passed through a fully connected layer with Softmax activation to predict probabilities $Y'$ over the three classes.

\vspace{-10pt}
\begin{align}
    Y' = \text{Softmax}(GW + b)
\end{align}


\subsection{Loss}
The model computes categorical cross-entropy loss $L$ between the predicted class probabilities and the correct class for the sample, as given below

\vspace{-10pt}
\begin{align}
    L(Y, Y') = - \sum Y \odot \log(Y')
\end{align}


\section{Experimental Setup}
\label{experimental_setup}

\subsection{Dataset}
The data, provided by the task organizers Patwa et al.  \shortcite{patwa2020sentimix}, comprises of tweets that employ the Hinglish mix. The data is in CONLL format, with each word being tagged as belonging to the English or Hindi language if applicable, else tagged as O if neither or EMT if an emoticon. Every tweet has a corresponding ID and sentiment label attached to it.

\subsection{Data Preprocessing}
We experimented with multiple steps for cleaning the tweets, including but not limited to lowering text case, expanding English contractions, replacing emoticons by their meanings in English, removing character repetitions, and removing hashtags, usernames and links. After cleaning, the data was tokenized and indexed by word and converted into embeddings. Features such as language labels for each word were converted to one hot vectors and concatenated to the word vectors (not present in final submission).

\subsection{Additional Details}
The 1D CNN layer has 200 filters, each of kernel size 8 with stride of 1. We used randomly initialized word embeddings of size 200. The model was optimized using Adam with a learning rate of $0.01$, \textit{beta1} as $0.9$, \textit{beta2} as $0.99$ and \textit{epsilon} as $1e-7$.

All the experiments were carried out on a single Nvidia GeForce RTX 2080 Ti GPU. The experiments were written in Python3, using the Keras framework \cite{chollet2015}. The code itself is hosted on GitHub under the MIT license\footnote{\href{https://github.com/IamAdiSri/hcms-semeval20}{Link to code repository.}}.

\subsection{Evaluation Metric}
For evaluation we simply use F1 as our metric, emulating the task leaderboard.


\section{Results and Analysis}


We used the official validation and test sets provided. While our contest results stand at 67.1\% F1, we have fine tuned our model since, and the latest scores for all experiments are shown in table \ref{tab:results}. (Contest results are visible on the leaderboard under the authors' aliases, \textit{the0ne} and \textit{talent404}.)\footnote{ \href{https://competitions.codalab.org/competitions/20654\#learn_the_details-results}{Link to contest leaderboard - authors' usernames are \textit{the0ne} and \textit{talent404}.}}

\begin{center}
    \resizebox{0.9\textwidth}{!}{
        \footnotesize
        \begin{tabular}{ |p{1cm}|p{5cm}|p{1.3cm}|p{1.3cm}|p{1.3cm}|p{1.3cm}|  }
            \hline
            \bf{S.No.}& \bf{Model Description}& \bf{Precision}& \bf{Recall}& \bf{Acc.}& \bf{F1}\\
            \hline
            1.& HCMS& 69.51& 68.01& 68.10& \bf{68.44}\\
            2.& BiLSTM + self-attention& 67.67& 67.44& 67.23& 67.49\\
            3.& HCMS w/o self-attention& 66.52& 66.75& 66.17& 66.56\\
            4.& fastText Classifier& 67.16& 65.74& 65.77& 66.26\\
            5.& BiLSTM w/o self-attention& 71.39& 65.02& 65.67& 66.18\\
            6.& SVM& 65.86& 67.20& 66.13& 65.97\\
            \hline
        \end{tabular}
    }
    \vspace{-5pt}
    \captionof{table}{Model descriptions and their results arranged in descending order of test F1 scores.}
    \label{tab:results}
\end{center}

The SVM \cite{svm} trained on n-grams of sizes 1, 2 and 3 returned an F1 score of 65.97\%. Given its ability to handle out-of-vocab words, fastText \cite{Joulin_2017} (also an n-gram based model) outperformed the SVM model, achieving 66.26\% F1.

Ablating the self-attention mechanism from the HCMS model left a vanilla 1D CNN with max pooling. Both, the LSTM and the CNN performed similarly to the n-gram based fastText model, but it is clear from the results that applying self-attention on top of features extracted by the respective models improved their performance significantly, providing credibility to our hypothesis of local and global features in the data. HCMS can be seen outperforming the models lacking self-attention by a significant \texttildelow1\%, giving a score of 68.44\%.




\begin{center}
    \resizebox{0.5\textwidth}{!}{
        \footnotesize
        \begin{tabular}{ |p{5cm}|p{1.5cm}|  }
            \hline
             \bf{Preprocessing Type}&  \bf{F1}\\
            \hline
                Emoji and contraction replacement& \bf{68.44}\\
                Only contraction replacement& 68.25\\
                Only emoji replacement& 68.18\\
                No Cleaning& 65.97\\
            \hline
        \end{tabular}
    }
    \vspace{-5pt}
    \captionof{table}{Preprocessing steps arranged in descending order of test F1 scores.}
    \label{tab:data_preprocessing}
\end{center}

It was observed that cleaning the data improved performance. Table \ref{tab:data_preprocessing} shows how specific forms of data preprocessing affect HCMS. Emoji replacement refers to the substitution of graphical emojis with their ASCII counterparts.\footnote{\href{https://pypi.org/project/emoji/}{Links to library used for emoji replacement.}} Contraction replacement refers to the substitution of English contractions like \textit{can't} or \textit{wouldn't} with their complete forms \textit{cannot} and \textit{would not} respectively. We can see in the table that using both gives the best performance.

\begin{figure}[!h]
\footnotesize
\centering

\begin{minipage}[t]{.44\textwidth}
\centering
\subcaption{Sentiment distribution over the corpus.}\label{man}
\end{minipage}%
\begin{minipage}[t]{.44\linewidth}
\centering
\subcaption{Language distribution over the corpus.}\label{woman}
\end{minipage}

\vspace{-15pt}
\includegraphics[trim=0cm 0cm 0cm 0cm, clip, width=0.8\textwidth]{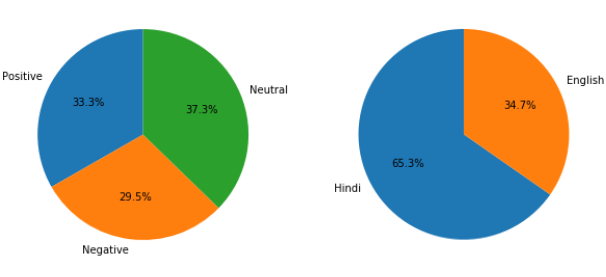}

\vspace{-10pt}
\caption[pie-charts]{Data distribution charts.\label{fig:pie-charts}}
\end{figure}

We also analysed the data and observed language distribution across the corpus. The numbers have been presented in figure \ref{fig:pie-charts}. As expected, the vocab is not equally split between the two languages.

In fact, there is greater presence of Hindi, and that too is Romanized and not in its native Devnagari script. In our experiments we tried to use a pretrained transliterator to de-Romanize the Hindi words, and then tried to employ pretrained word-embeddings, models and sentiment lexicon - due to the noisy Devnagari-to-Roman transliteration by the authors of the tweets however, our reverse transliteration failed and most words did not map into their respective embedding spaces.\footnote{\href{https://github.com/libindic/indic-trans}{Links to Indictrans Transliterator}, \href{https://fasttext.cc/docs/en/aligned-vectors.html}{aligned fastText embeddings}, \href{https://github.com/google-research/bert/blob/master/multilingual.md}{multilingual-BERT} and \href{https://www.kaggle.com/rtatman/sentiment-lexicons-for-81-languages}{sentiment lexicon.}} The results of those experiments are presented in table \ref{tab:failed_exp} (note that the other steps of the data cleaning process were preserved).

\begin{center}
    \resizebox{0.7\textwidth}{!}{
        \footnotesize
        \begin{tabular}{ |p{8cm}|p{1.5cm}|  }
            \hline
             \bf{Experiment}&  \bf{F1}\\
            \hline
                Reverse transliteration + Aligned fastText &55.9\\
                Reverse transliteration + Multilingual BERT &52.3\\
                Reverse transliteration + Sentiment annotation &52.7\\
            \hline
        \end{tabular}
    }
    \vspace{-5pt}
    \captionof{table}{Failed experiments due to noise in the data.}
    \label{tab:failed_exp}
\end{center}

In our experiments, another noteworthy occurrence was that even though some models showed high validation scores, they were outperformed on the test set by models that produced lower validation score in comparison. This was a recurring theme in our A-B testing over validation and test data, and we feel that this can again be attributed to the noisy nature of code-mixed language.

\section{Conclusion}

In this paper we have discussed a way to perform sentiment analysis on code-mixed Hindi-English data, and we feel that the system is versatile enough to be applied to any mix of languages, especially when resources for the mixture are unavailable or hard to find. We have also looked at ways that such data can be preprocessed for training, and how a multilevel neural architecture is able to extract context from text.

While it is encouraging to see that our model performs consistently, there are a couple of caveats that need to be addressed - even though the lack of pretraining, a low resource setting, and noise in data seem surmountable hurdles to the task, it is evident that the most promising avenues for improvement lie in exploring data cleaning/normalization and transfer learning. Further analysis of the present methodology could also reveal the kind of transfer learning required - whether it be pretrained word embeddings, large scale language modeling or perhaps a better transliteration pipeline to name a few. To do so would require larger quantities of data and the application of the system to other linguistic tasks of settings similar to that of Sentimix. All of these steps qualify within the scope of future work, and committing to them will reward us with more confidence in the hypotheses we propose in this paper, providing further proof of the system's robustness and effectiveness. We hope to undertake efforts for the same soon.



\bibliographystyle{coling}
\bibliography{semeval2020}

\end{document}